\documentclass[letterpaper]{article} 
\usepackage{aaai2027}
\usepackage{times}
\usepackage{helvet}
\usepackage{courier}
\usepackage[utf8]{inputenc}
\usepackage{amsmath}
\usepackage{amsfonts}
\usepackage{amssymb}
\usepackage{graphicx}
\usepackage[hyphens]{url}
\usepackage{caption}
\usepackage{booktabs}
\usepackage{multirow}
\usepackage{array}
\usepackage{url}
\usepackage{hyperref}   
\nocopyright               
\title{Perforated Backpropagation: A Neuroscience Inspired Extension to Artificial Neural Networks}

\author{
    Rorry Brenner\textsuperscript{\rm 1,2},
    Laurent Itti\textsuperscript{\rm 1}
}
\affiliations{
    \textsuperscript{\rm 1} Computer Science and Neuroscience, University of Southern California\\
    \textsuperscript{\rm 2} Perforated AI Inc.\\
    rorry@perforatedai.com, itti@usc.edu
}

\begin{document}

\maketitle

\begin{abstract}
The neurons of artificial neural networks were originally invented when much less was known about biological neurons than is known today. Our work explores a modification to the core neuron unit to make it more parallel to a biological neuron. The modification is made with the knowledge that biological dendrites are not simply passive activation funnels, but also compute complex non-linear functions as they transmit activation to the cell body. The paper explores a novel system of ``perforated'' backpropagation empowering the artificial neurons of deep neural networks to achieve better performance coding for the same features they coded for in the original architecture. After an initial network training phase, additional ``dendrite'' nodes are added to the network and separately trained with a different objective: to correlate their output with the remaining error of the original neurons. The trained dendrites are then frozen, and the original neurons are further trained, now taking into account the additional error signals provided by the dendrites. The cycle of training the original neurons and then adding and training dendrites can be repeated several times until satisfactory performance is achieved. Our algorithm was successfully added to modern state-of-the-art PyTorch networks across multiple domains, improving upon original accuracies and allowing for significant model compression without a loss in accuracy.
\end{abstract}


\section{Introduction}

Advances are continuously being worked on in deep learning. After the large resurgence of neural nets for computer vision with the success of AlexNet~\cite{Krizhevsky2012} many tweaks have been made to the system of connections between layers. From Google's Inception Modules~\cite{Szegedy2015} to Microsoft's ResNets~\cite{He2016} to modern image transformers~\cite{Parmar2018} the results on the ImageNet object detection challenge~\cite{Russakovsky2015} are improved every year.

The first mathematical model of an artificial neuron was invented in 1943~\cite{McCulloch1943} based on 1943 neuroscience. This idea was brought into artificial intelligence via the perceptron in 1958~\cite{Rosenblatt1958}. Backpropagation through layered networks was added in 1970~\cite{Linnainmaa1970}. Since then, computer scientists have innovated new activation functions, optimizers, schedulers, and architectures, all while the hardware exponentially improved to bring us to where we are today. However, the neuron unit and the core backpropagation algorithm based on gradient descent have remained unchanged since their original invention. This paper explores the modern neuroscience understanding of active dendrites~\cite{Major2013} and how such a concept could be added to an artificial neural network.

When exploring how to best instantiate this method, two restrictions were set. First, when active artificial dendrites are created, they must be outside of the network despite being able to affect individual neurons within the network. Second, something about them must operate differently than traditional artificial neurons. The Perforated Backpropagation paradigm achieves both goals. Critical to note, Perforated Backpropagation is an addition, and not a replacement. This means that it is not a novel architecture which would replace the work others have done, e.g., using this instead of a transformer model or a ResNet. Perforated Backpropagation is a plug-in that functioning systems can quickly add, rather than a method that would require researchers to replace any other component of what makes their systems unique and successful or perform any major overhauls of their processes. By adding dendrites to the neurons of state-of-the-art systems, Perforated Backpropagation is able to push the state of the art even further by keeping the same overall architectures but making selected neurons multi-node modules, which perform better at their original roles.

\section{Background}

\subsection{Biological Background: Active Dendrites}

The human brain is an amazing computational machine. Its visual system has been said to be the only evidence that generalized object recognition is possible. This is why artificial neurons follow its design in a computational parallel. However, the original artificial neuron is a significantly reduced version of a biological neuron. This artificial neuron is still highly functional because to capture all of the functional capabilities of a biological neuron, an artificial neuron does not need to model every single atom. The question becomes, not only what functions are missing in artificial neural networks that biological networks perform, but also what functions are essential for their aptitude. The function to inspire this work is that of active dendrites.

In modern state-of-the-art artificial neural networks, dendrites are nonexistent~\cite{Krizhevsky2012, Szegedy2015, Ciregan2012, Kubilius2016}. Presynaptic neurons form synapses directly to the cell bodies of postsynaptic neurons with connection weight as their only parameter~\cite{Widrow1990}. These algorithms are not at a place where they can compete with the human visual system. Artificial networks differ in many ways from those of biology, but one certain advantage the human visual system has is the increased computational power within a single neuron by way of their dendrites.

The simplest function dendrites perform is physically being the structure on which most synapses form to collect presynaptic activation and transport it to the soma~\cite{Kandel2000}. Even this most basic function still behaves in a way that must be considered. Dendrites are not perfect cables, they are leaky. Voltage attenuates significantly if it travels down the dendrite with no further external influence~\cite{Stuart1998}. The axial resistance is lower towards the soma so voltage will flow towards the cell body before leaking out through the membrane. Collecting voltage for the cell is only a fraction of their complete ability. Passive summation over dendritic voltages plays an important role. Coinciding peak voltages cause a greater surge, more likely to cause a spike, at the soma. This allows the neuron a type of temporal calculation of synapse firing order which might only peak if inputs are in a proximal-to-distal order~\cite{Rall1964, Koch1983}. Another important calculation dendrites provide, and the one we will discuss adding to artificial neurons, is their active ability to generate spikes. The exact details of this ability are still disputed by some scientists. A paper by Major et al. shows a framework in which conflicting results can all fit~\cite{Major2013}.

NMDA is the dominant spike type in dendrites. NMDA allows both thresholded and graded electrogenesis. Depending on glutamate concentration from synaptic firing, NMDAR conductance can be up-stable, down-stable, or unstable. Unstable is the condition which allows for a thresholded response. NMDAR conductance has a unique N-shaped I-V curve. This N shape can have one early intercept, three intercepts, or one later intercept. At low glutamate concentrations the trough of the curve representing current flow is completely above the axis meaning current flows outward at typical biological membrane potentials. As the concentration increases the trough gets deeper until it hits a point where it dips below zero. This causes the curve to cross the axis at three points. The middle crossing is an unstable state for voltage. If the instantaneous voltage is below this intercept voltage will flow outward, lowering the membrane potential until it reaches the first intercept, which is a stable state. If voltage is allowed to surpass this point through presynaptic firing the current is now inward, causing the membrane to further increase up to the third and final intercept, creating a voltage spike.

An NMDAR I-V curve being based on glutamate concentration causes an interesting effect. Unlike the voltage dependent spiking of an action potential, NMDA spikes are still voltage dependent, but the exact voltage can change. It is still all-or-none, in the sense that there is a voltage threshold above which the voltage will spike to a higher value. This threshold, however, does not exist at low glutamate levels, slowly lowers as concentration increases, and then ceases to exist again at high concentrations~\cite{Major2013}. The threshold and amplitude of the spike both increase from distal to proximal dendritic locations in correlation with the respective input conductance. Clearly, NMDA receptors are already dynamic on their own, but this nonlinear function only gets more complex when considering a system with multiple connections. All of this discussed functionality is missing from current artificial neural network models. This significantly impedes the potential of artificial neural network performance, and some neurobiological research supports the idea that dendrites, and not neurons, are the fundamental functional unit of the brain~\cite{Branco2010}.

\begin{figure}[t]
\centering
\includegraphics[width=\columnwidth]{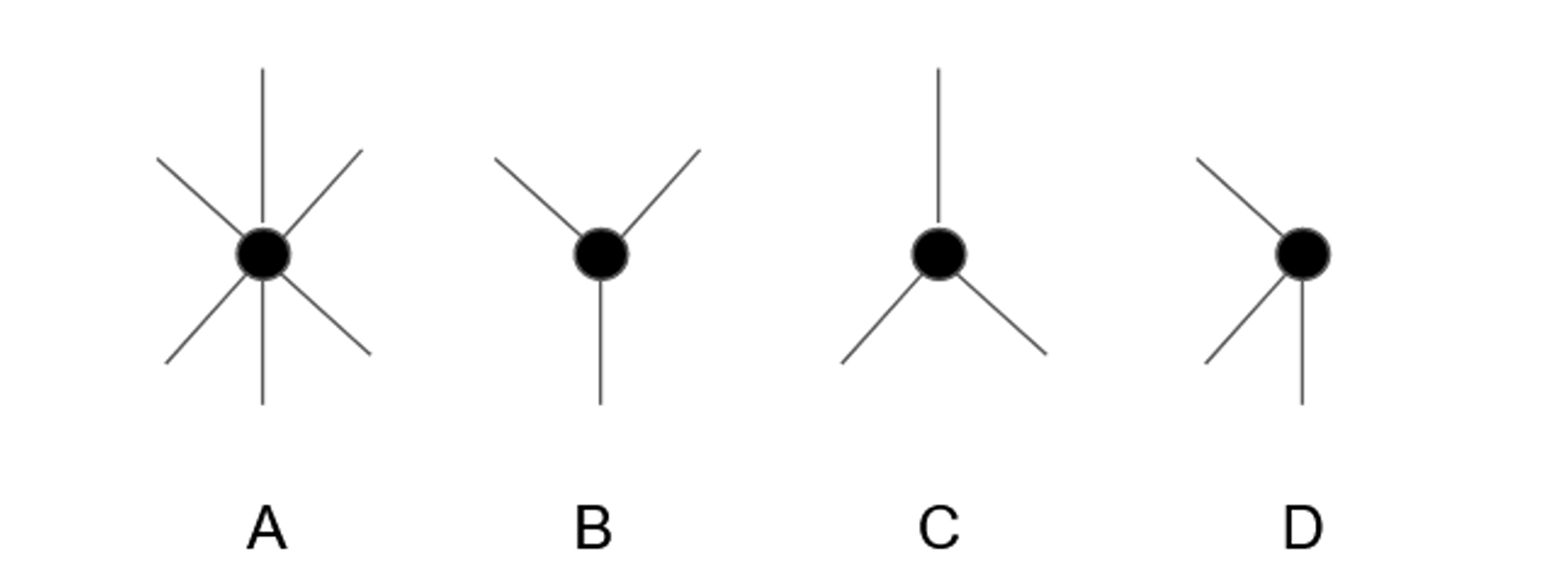}
\caption{Four examples of connected lines. B and C are y-vertices, D is an arrow-vertex, and A is six lines coming together at a point.}
\label{fig:vertices}
\end{figure}

This ability of biological dendrites allows individual biological neurons to code for features that would require multiple neurons in more than one layer in an artificial network. An easy to visualize example comes from the non-accidental properties of Biederman's geon theory of shape perception~\cite{Biederman}. Figure~\ref{fig:vertices} shows an example of four vertices. B and C are y-vertices, while D is an arrow-vertex and A is 6 lines converging on a point. Our visual system naturally associates distinct partial three-dimensional shapes to some of these: B and C might be perceived as a corner facing the observer, while D is a corner facing away, and A is an assortment of lines. Individual neuronal recordings have shown that, in biological visual systems, neurons are sensitive to these types of non-accidental properties when looking at objects~\cite{Kayaert2003}. A single artificial neuron could not code for a feature like this because if the lines provide positive input, B and C could not cause high activation while A and D do not. The motivation to add artificial dendrites to the neurons of an artificial network is to allow individual neurons to learn more complex features such as these.

As artificial neurons do not replicate the exact function of biological neurons, the dendrites of Perforated Backpropagation do not replicate the exact function of NMDA spikes. But similar in function to biological dendrites, this work explores if better systems can be created by granting individual neurons the ability to make more complex decisions by adding nonlinear processing between each neuron and the presynaptic neurons from the previous layer. However, this nonlinear processing is added in a different way than the spline method of the recently published paper on Kolmogorov-Arnold Networks~\cite{Liu2024}. To simplify this system and allow it to be compatible with modern deep learning frameworks, here additional tensor processing layers are used as a proxy for artificial computation-enabled dendrites. However, a method is devised to ensure that this particular subset of layers is outside of the network, by adjusting the way they interface with traditional backpropagation of modern deep learning systems. This way, we propose a new network design and information processing paradigm that is distinct from adding more neurons and connections to a standard deep neural network.

\subsection{Computational Background: Cascade Correlation}

One of the restrictions set for this research was ensuring dendrite nodes behaved or learned differently than neuron nodes. To that end, a starting point was chosen from an earlier algorithm which also instantiates a network that has nodes which operate differently from each other. Cascade Correlation~\cite{Fahlman1989} is a system developed to improve single-layer networks. It utilizes artificial neurogenesis to continuously improve error by adding new hidden nodes, which we refer to as dendrites, between the input and output layers. Learning takes place by alternating between two phases: typical perceptron learning, and dendrite learning. During perceptron learning, each output node learns as a standard perceptron to minimize its error based on its inputs. During dendrite learning, first, the weights of the network are frozen. Next a set of candidate dendrites are created which are not a part of the network yet. They are given inputs from the same input nodes as the perceptrons of the output layer, but they form no output connections. To update their input weights, they use a system of localized learning with an objective based on the covariance between their own output and the residual error of the neuron nodes. When given a test point, this objective is highest when the candidate's output rises and falls together with the network's error, and lowest when the two are unrelated or move oppositely. With this process new dendrites learn to maximize the correlation between their output and the error of the network.

\begin{figure}[t]
\centering
\includegraphics[width=\columnwidth]{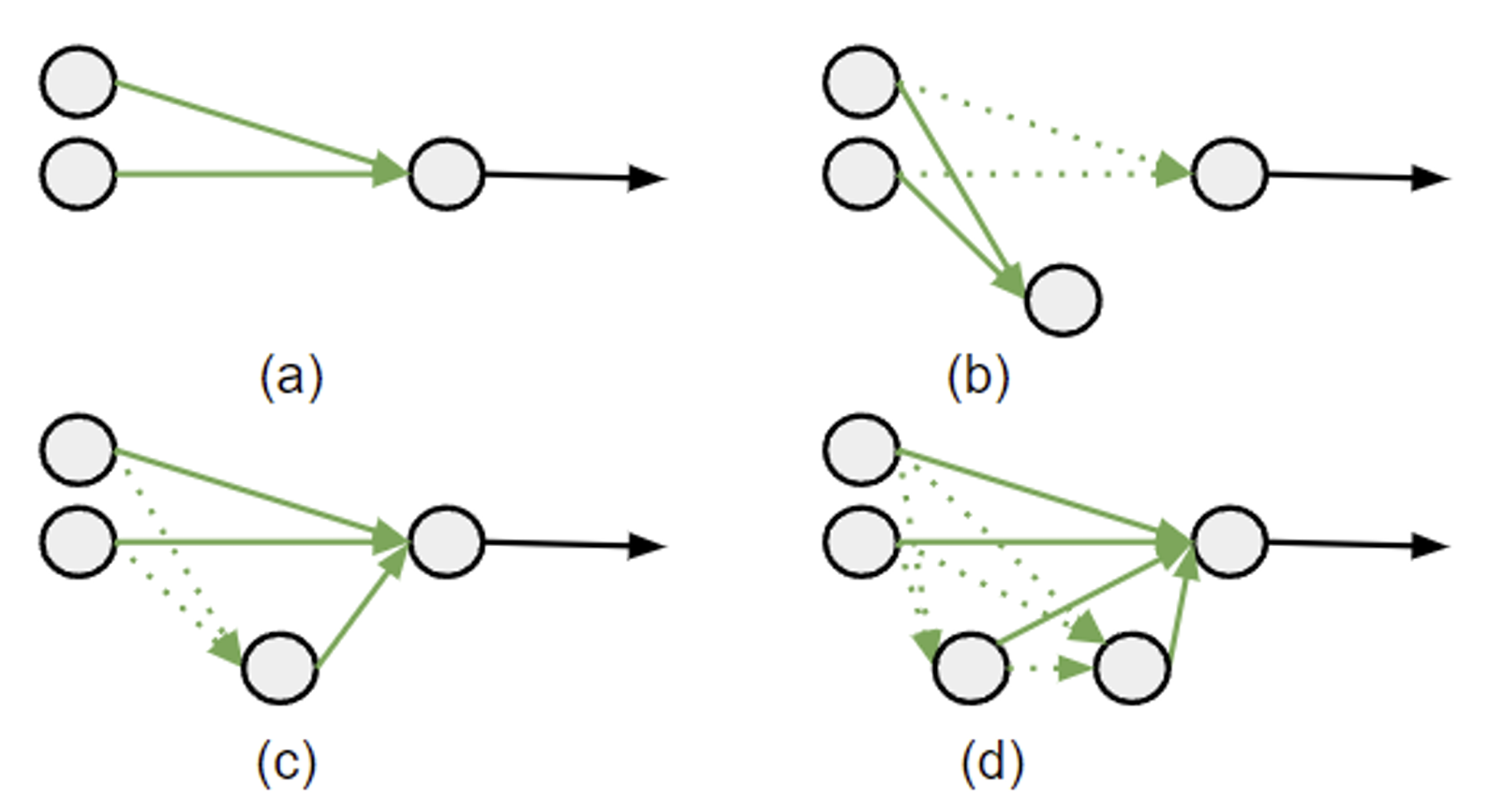}
\caption{Stages of Cascade Correlation learning are shown. In [a] a perceptron is shown with 2 inputs and 1 output. In the first phase this perceptron is trained to maximize classification of the input data. Next in [b] the weights from the input to the network are frozen, signified by dotted lines, and the new dendrite learns weights which maximize its correlation with the perceptron's error. [c] shows the network in the next phase with the new dendrite added after having its weights frozen and the perceptron learning to modify its input weights as normal with the one additional input from the new node. In [d] an additional dendrite has been added with connections from both the input and the other created dendrite.}
\label{fig:cascade_stages}
\end{figure}

\begin{figure}[t]
\centering
\includegraphics[width=\columnwidth]{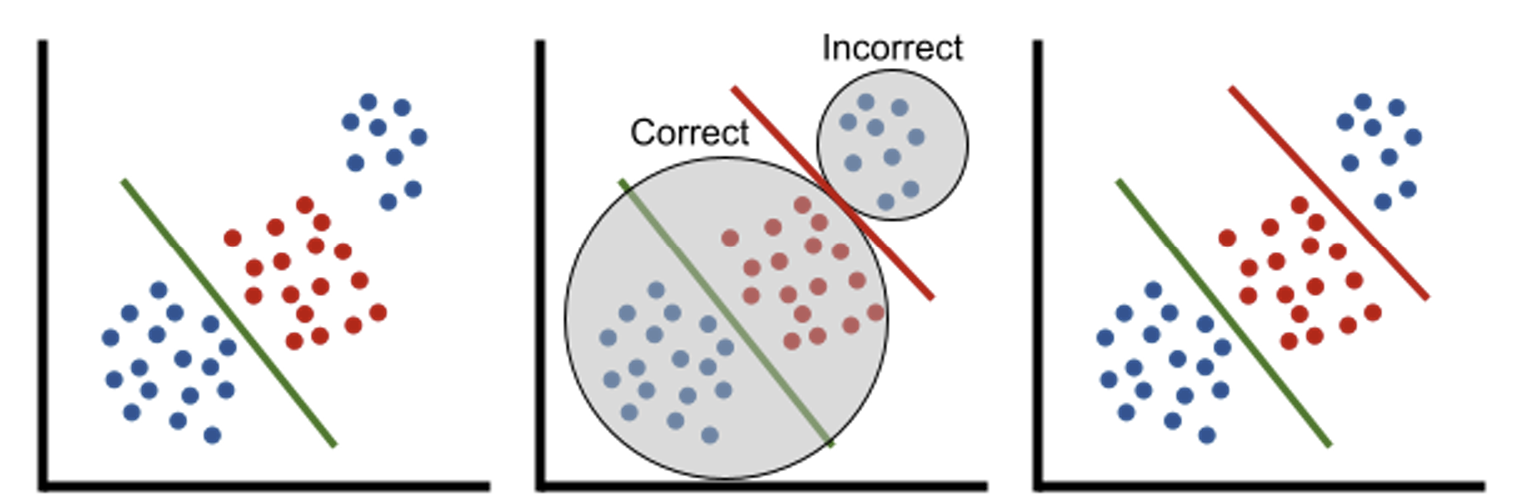}
\caption{These graphs show the linear classification capabilities of a Cascade Correlation network. On the left a single perceptron can only create a single classification line. For simplicity, this line has been drawn to perfectly classify the blue data on the left and all the red data, while getting its classification wrong on all the blue data on the right. The center image shows how the dendrite learns. Rather than classifying the data, it learns to classify where the perceptron is wrong. If there were more than one output node it would learn to classify where all of the outputs make the most mistakes. On the right the red line represents the classification line of the new node. The output perceptron now has not only the x and y coordinates as input but also an input representing if those points are above or below the red line. It can now modify its three weights to correctly classify all of the data.}
\label{fig:cascade_classification}
\end{figure}

Once training is complete the candidate dendrite which achieved the best correlation score is added to the network. This is done first by fixing its input weights, which are never changed again, forming a permanent snapshot of the network's error at the timepoint it was added. Next the candidate dendrite forms output connections to each of the output nodes. Then the dendrite training phase is over, switching back to perceptron learning where each output node treats their new input connection identical to their connections directly from the input. The only additional complexity is that during the next phase when new candidate dendrites are added they also form input connections from all previously created dendrite nodes. Figures~\ref{fig:cascade_stages} and~\ref{fig:cascade_classification} show visuals to explain the architecture itself, and what the architecture can do. For simplicity they both use two inputs to represent x and y coordinates, making the perceptron a linear classifier.

\section{Related Work}

The concept of adding artificial dendrites to machine learning models of artificial neurons has been around since 2003~\cite{Ritter2003}. Early work focused on morphological perceptron models while modern works also leverage traditional artificial neuron models. Traditional neural networks use a weighted sum leading into a nonlinearity. Morphological perceptron models instead use lattice algebra including maximum and minimum functions. Works related to artificial neuron models with dendrites mostly focus on shallow networks with only a single layer of neurons and any additional layers composed of dendrites. Because they use shallow networks, they tend to focus on efficiency rather than accuracy. A review paper from 2021~\cite{Chavlis2021} found that none of the dendrite models they reviewed achieved state-of-the-art accuracy on the datasets they were applied to. However, they did find the models achieved better performance compared to traditional networks with the same number of parameters.

From the morphological subset one group of authors~\cite{Sossa2014} introduces the addition of dendrites to morphological perceptrons which can train dendrites to perfectly classify any training dataset. The algorithm instantiates a neuron with dendrites for a set of input data by iteratively creating hypercubes which encapsulate possible input patterns. First a hypercube is generated which encapsulates all the data. Then if the training data inside a hypercube is entirely within a single class the hypercube is saved. If the training data is not entirely within a single class, the hypercube is iteratively split into smaller hypercubes. Once this process is complete the neuron is given a set of dendritic branches corresponding to each hypercube where the branches have frozen weights such that a particular branch is only active if a datapoint is within its associated hypercube. The authors showed this algorithm could perform well on basic pattern recognition and even be applied to image processing tasks outperforming simple multi-layer perceptrons. Similar work extended this concept to use hyperellipsoids in place of hyperboxes, fitting each one's centroid and covariance with k-means++~\cite{Arce2018}.

Research has also been performed which sets up architectures that are inspired by dendrites, but follows traditional neural network theory by using standard backpropagation for training. One set of authors~\cite{Li2020} was able to show this architecture could achieve improved performance with significantly reduced power consumption from hardware. Another paper~\cite{Meir2023} explored a dendritic structure applied to CIFAR-10~\cite{Krizhevsky2009} which outperformed the larger LeNet~\cite{LeCun1998}. However, these experiments were performed comparing the methods to other small models and not state-of-the-art architectures.

In addition, artificial dendrite research has been performed which uses traditional neurons but strays from traditional neural network training by not using backpropagation as its training paradigm~\cite{Gao2018}. Instead, the authors ran experiments using six learning algorithms: biogeography-based optimization, particle swarm optimization, genetic algorithm, ant colony optimization, evolutionary strategy, and population-based incremental learning. While the original authors only compared this method to multi-layer perceptrons, further research~\cite{Egrioglu2022} showed this algorithm to be the best machine learning method for a time series forecasting problem using the particle swarm method, though it did not outperform algorithms which did not use machine learning. Another group of researchers~\cite{Guerguiev2017} used the dendrites themselves to develop an alternative error signaling that allows error to be transmitted from the output layer to hidden layers with a more biological method than traditional backpropagation. This system used ``basal'' compartments for forward processing of input data and ``apical'' compartments for backwards processing of output error.

A key difference between our method and the discussed research is that other dendrite methods fundamentally change the network or training paradigm. For a modern machine learning researcher to switch from a standard neural network to a neural network with dendrites these methods would at a minimum require adjustment of the definitions of their models to include additional sparse layers, and in most cases require a fundamental change from the training process they currently use to switch to the new paradigm. One of the advantages of Perforated Backpropagation is that it is not a shift away from the norm that requires significant changes to modern methods. Instead, it is a wrapper that can be applied to the modules of standard deep learning pipelines that requires no adjustment to model definitions, training parameters, or datasets. With Perforated Backpropagation the model is wrapped in a way that allows the existing neurons within tensor based deep neural networks to be optimized with dendrites. This direct compatibility lets Perforated Backpropagation combine with other modern advances in architecture and training paradigms rather than compete with them. The experiments in this initial paper are confined to models that fit on a single GPU, while ongoing work being prepared for publication applies the same wrapper to substantially larger architectures and to real-world deployed systems. To our knowledge, this makes Perforated Backpropagation the first dendritic method to scale beyond academic settings. 

\section{Perforated Backpropagation}

Similar to the process in Cascade Correlation, during Perforated Backpropagation the network is first fully trained with traditional backpropagation until no more improvements to training error are made with additional epochs through the training data. At this point every neuron in the network is given a set of associated candidate dendrites, which behave in the same way as the candidate nodes of a Cascade Correlation system. All the math from Cascade Correlation learning transfers over with only a minor modification. The difference to consider is that the original Cascade Correlation system was only created to work with perceptrons, i.e. networks where the original architecture had no hidden layers at all. To apply this methodology to deep learning a change that must be made is adding dendrites to deeper neurons in the hidden layers of the network. The problem of how their dendrites learn is solved by relying on the error method already in use, backpropagation. However, the key contribution of this paper is the development of the novel ``Perforated'' Backpropagation paradigm.

During Perforated Backpropagation, error from the output is propagated through the network, but the calculations involved in the gradient descent equations only calculate error terms for neurons and only factor weights and output values from neuron-to-neuron connections, completely ignoring connections going through dendrites. However, the error gradient calculated at each neuron is not only used to update its weights during normal learning, but during dendrite learning the dendrites learn to maximally correlate their output activation with the backpropagated error of the single neuron they are connected to. This training will continue until no neuron has a dendrite which is still improving its correlation. Perforated Backpropagation allows the network to both stay the same and be added to. By removing the dendrites from the gradient descent equations, the model continues to have the same network architecture, and autograd graph, as defined by the error communication. This allows each neuron in a network to continue to code for the same feature, and have the same role, as it did in the original network. However, each neuron is also empowered to be better at coding for that feature by perforating its calculated error to one or more dendrites that can learn to identify patterns where the neuron is making mistakes and then adjust the weight of its connection to those dendrites to make better decisions about these outliers.

Traditional backpropagation through a multi-layer perceptron uses the equation from~\cite{Norvig2003} with the following definitions. $\Delta_j$ is the error associated with neuron node $j$. $g'(in_j)$ is the derivative of the activation function $g$ applied to the total input to neuron node $j$. Finally, $\sum W_{i,j}\Delta_i$ is the sum of each weight between postsynaptic neuron node $i$ and presynaptic neuron node $j$ multiplied by the error associated with neuron node $i$.

\begin{equation}
\Delta_j = g'(in_j) \sum_i W_{i,j}\Delta_i
\label{eq:traditional_bp}
\end{equation}

For Perforated Backpropagation, this equation is modified such that postsynaptic nodes are split into postsynaptic neuron nodes and postsynaptic dendrites. With postsynaptic neuron nodes designated as $i$, and post synaptic dendrites designated as $k$, the equation is split across the summation with the dendrite error terms zeroed out during the backpropagation process.

\begin{equation}
\Delta_j = g'(in_j)\left( \sum_i W_{i,j}\Delta_i + 0 \times \sum_k W_{k,j}\Delta_k \right)
\label{eq:perforated_bp}
\end{equation}

Additionally, with Perforated Backpropagation $\Delta_k$ can be calculated differently. For these first experiments, $\Delta_k$ was chosen with a slight modification to the original Cascade Correlation formula which is originally taken from the formula for covariance of two variables. This modification was made because the original formula takes multiple output neurons into account, as well as expecting significantly smaller datasets. In our case the following formula is applied during each mini-batch with the following definitions. $g(in_k)$ is the activation function of the dendrite applied to the sum of its input. $\bar{g}(in_k)$ is a running average over previous activations of the dendrite. Similarly, $\bar{\Delta}_i$ is a running average of the errors of the associated neuron node.

\begin{equation}
\Delta_k = (g(in_k) - \bar{g}(in_k))(\Delta_i - \bar{\Delta}_i)
\label{eq:dendrite_error}
\end{equation}

The weight update rule for input connections remains largely the same as the original paper with parallel changes given the updated $\Delta_k$. The new symbol $\sigma$ in this formula is the sign of the average correlation between the dendrite's output and the associated neuron node's output.

\begin{equation}
\frac{\partial\Delta_k}{\partial w_j} = \sigma(\Delta_i - \bar{\Delta}_i) g'(in_k)
\label{eq:weight_update}
\end{equation}

Rather than choosing a single candidate node to add to the network, as in Cascade Correlation, every selected neuron adds its associated candidate dendrite with the highest correlation to its error. After adding the candidate dendrites, all their input weights are locked. At this point each neuron has the same inputs as before, but with the addition of one single input from its new associated dendrite.

With dendrites added, the network once again performs backpropagation learning. However, the error backpropagation is still performed in exactly the same way, which means only through the neurons of the network. The connection weights from the dendrites are modified together with the connections from the other presynaptic neurons in the network, but no error is propagated through them. Once again with backpropagation learning the network will reach a learning asymptote. When this occurs another set of dendrites is added in the same manner. The one difference being that new dendrites for a neuron also receive input from all previous dendrites created for that individual neuron, as is done in Cascade Correlation. Adding these artificial dendrites to neurons of an established pre-trained network allows the neurons to continue to code for the same feature they were coding for in the original design. However, each additional dendrite allows every neuron in the architecture to reduce their error at coding for that feature.

\begin{figure}[t]
\centering
\includegraphics[width=\columnwidth]{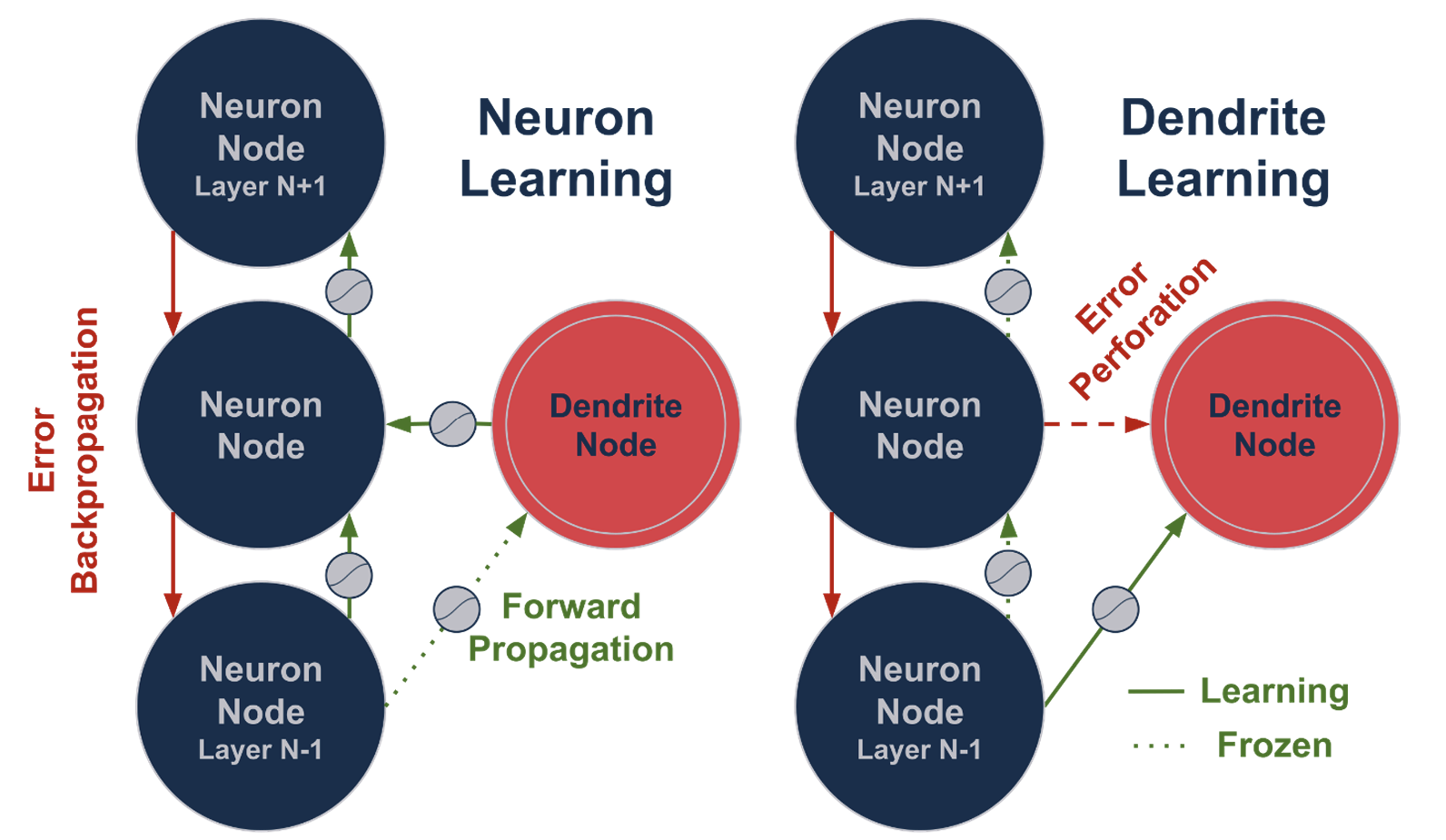}
\caption{The left image shows the process of neuron learning. Green lines show weights that are being updated. Red lines show the error being backpropagated through the layers. The dotted green line shows weights that are currently frozen and not being adjusted. The right image shows dendrite learning. In this case the dashed red line shows error that is being perforated out of the neuron to the dendrite which then learns its connection weights from the neurons in layer N-1.}
\label{fig:perforated_process}
\end{figure}

This achieves the two goals of the project. First the dendrites are outside the network. This is because, during backpropagation, the learning error is not passed through them. Gradient descent from the output error only flows through the neuron nodes in the network. This also means while the forward computation graph is altered, the backwards computation graph remains the same.  Retention of the original neuron-only computation graph is also why we define these multi-node components as a single neuron module. Additionally, the dendrites are outside the network because the original network layers are fully connected but each dendrite only connects to a single neuron postsynaptically. They are still able to affect individual neurons because, in the forward pass, their output connections are formed to their single associated neuron in addition to any other dendrites connected to the same neuron which were created after them. Second, they are different than the neurons in the way they learn. Rather than learning through gradient descent, they use the Cascade Correlation learning system but use the backpropagated error received by their associated neuron rather than the perceptron output error a traditional Cascade Correlation node would have used. A visualization of this process can be seen in Figure~\ref{fig:perforated_process}.

Perforated Backpropagation works not only for multi-layer perceptrons, but also convolutional methods. The behavior is the same where neurons are convolved around their input tensors with the dendrites being convolved at the same time. This means every neuron kernel only adds one single input connection per dendrite kernel it adds. That is, with an N$\times$N kernel, the postsynaptic neuron will have N$\times$N+1 input connections with a single dendrite.

\section{Results}

\subsection{Baseline test}

All experiments were performed with the expected training, validation, and test splits of the datasets. Calculating the scores represented in each graph was performed in the following way. As a network switches back and forth between training neurons and training dendrites there are a set of epochs and scores associated with each architecture. The training epochs associated with one architecture are defined as a ``cycle''. The validation score of a cycle is the highest validation score calculated during that cycle. However, the test score of a cycle is not the highest test score calculated. Instead, the cycle's test score is the test score that was calculated during the same epoch where the highest validation score was calculated. Additionally, the max test score of an experiment is defined as the test score calculated during the epoch where the max validation score was calculated. In each experiment training would take place with these pairs of values being saved until an architecture was created where the highest validation score of that cycle did not improve upon the previous one. This happens in every experiment as dendrites will always eventually begin to overfit the training data, so early stopping must be performed once this point is reached. Similarly, early stopping is performed during each neuron training cycle by loading the best set of weights of that cycle before adding dendrites for the next dendrite cycle.

\begin{figure}[t]
\centering
\includegraphics[width=\columnwidth]{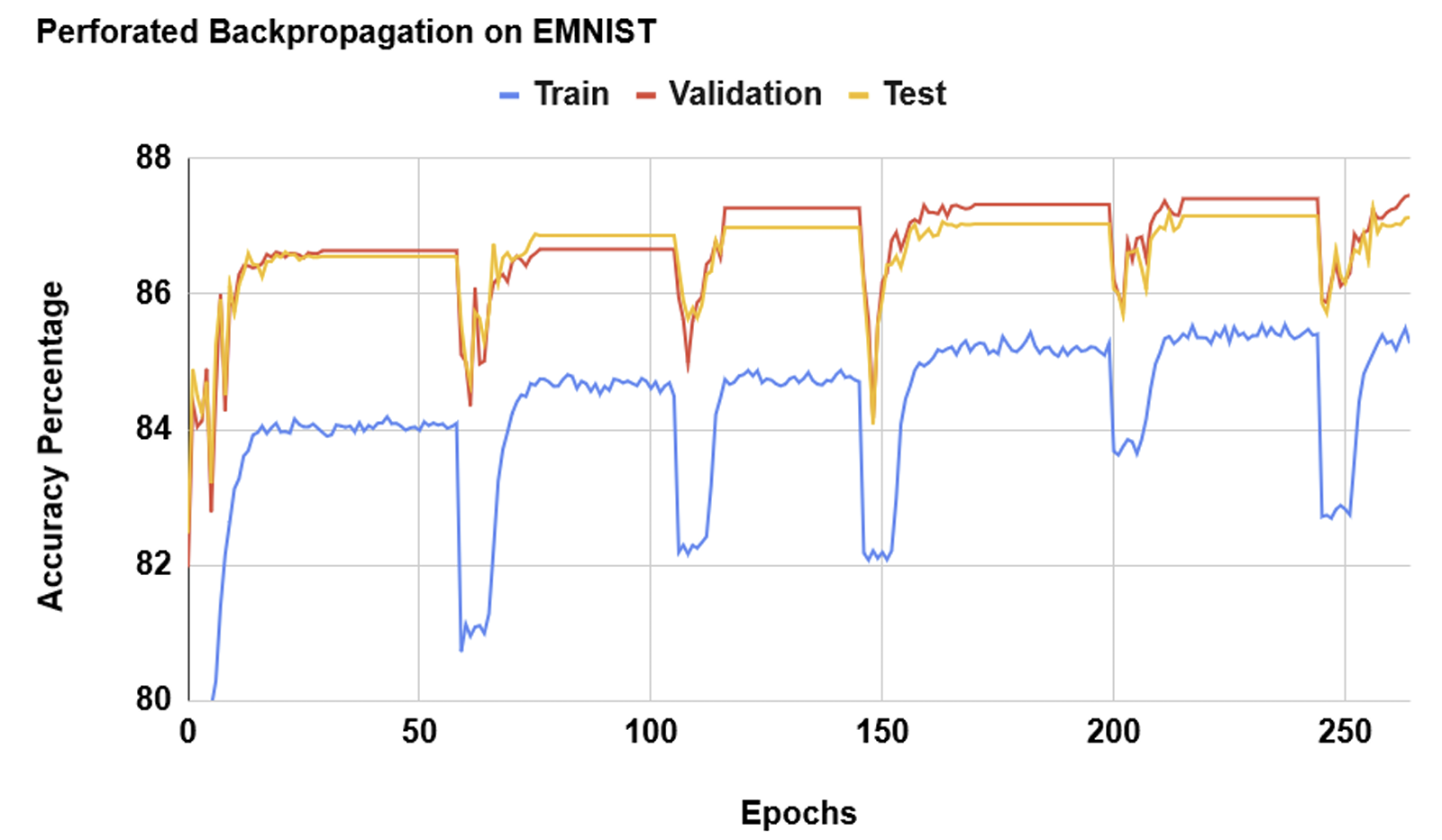}
\caption{Train, Validation, and Test scores for each epoch of an experiment of Perforated Backpropagation on EMNIST. Epochs in the graph where Test and Validation scores are flatlined over multiple epochs are dendrite training cycles where the neuron weights are frozen. Training scores of dendrite cycles still fluctuate because of random shuffling of the dataset. Each dip after a dendrite cycle occurs as the network adjusts to the new connections. Graph shows Perforated Backpropagation allows scores to go up each cycle after training reaches threshold.}
\label{fig:emnist_scores}
\end{figure}

To show a proof of concept, experiments were run to optimize a simple training pipeline. The pipeline chosen was the MNIST~\cite{Deng2012} example from the official PyTorch GitHub repository~\cite{PyTorch2024}. Because this example started at near perfect accuracy, the MNIST dataset was replaced with the EMNIST Balanced dataset~\cite{Cohen2017} which contains the same format of images with letters in addition to the original numbers. The network trained has two convolutional layers of widths 32 and 64 leading into two fully connected layers of widths 9216 and 47. All layers use a ReLU activation function, there is a max pooling layer after the second convolution layer, and each fully connected layer is preceded by a dropout layer~\cite{Srivastava2014}. Training was performed over the EMNIST images with a new cycle beginning when no improvement was seen for 25 epochs in a row. A graph of the exact scores of one of these experiments can be seen in Figure~\ref{fig:emnist_scores}.

\begin{figure}[t]
\centering
\includegraphics[width=\columnwidth]{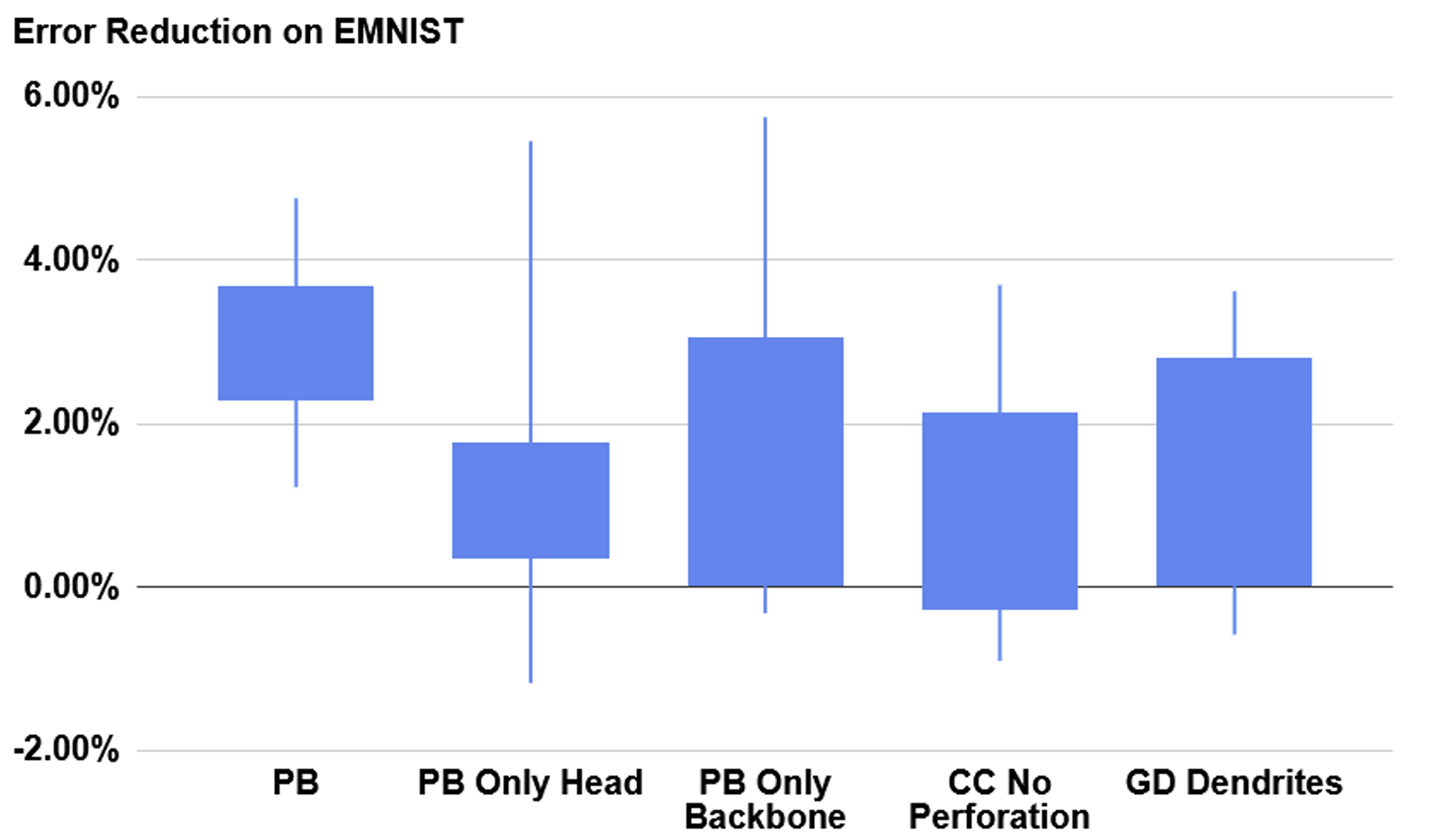}
\caption{Candlestick charts of test error reduction over seven experiments of each type performed on the EMNIST dataset comparing the test score of the original cycle with zero dendrites to the max test score. PB -- Perforated Backpropagation. PB Only Head -- Only adding dendrites to the output layer. PB Only Backbone -- Adding dendrites to all layers excluding the output layer. CC No Perforation -- Performing Cascade Correlation learning on each layer but also allowing the backpropagation error signal to flow through the new nodes. GD dendrites -- Stopping the error signal from flowing through dendrites, but training dendrite weights with gradient descent rather than Cascade Correlation learning.}
\label{fig:ablation_study}
\end{figure}

Additional tests were performed to analyze the importance of each of the components of Perforated Backpropagation. To minimize the impact of random initializations changing the starting scores, Figure~\ref{fig:ablation_study} shows values for percentage error reduction between the original cycle to the final cycle. For example, the experiment from Figure~\ref{fig:emnist_scores} has a first cycle test accuracy of 86.55\% and a final cycle test accuracy of 87.13\%, which means the remaining error was reduced by 4.3\%. Each experiment was performed seven times. Candlestick values in all charts are minimum, quartile one, quartile three, and maximum.

\subsection{State-of-the-Art Accuracy Tests}

The next experiments performed were on architectures which were state-of-the-art systems in various fields at the times they were published. In each case, the Papers with Code website~\cite{PapersWithCode2023} was referenced to find networks to improve. This website is a database of datasets. Each dataset contains a list of papers which have been published which utilized that dataset in their experiment. Links to the open-source repositories are included when code is available. Due to hardware restrictions, datasets were selected where the top systems could be trained with published settings on a single GPU. All experiments were run on an Ubuntu workstation with a NVIDIA GeForce RTX 2080 GPU and an AMD 9 3900X CPU. In addition, models were chosen that stated they achieved state-of-the-art performance without additional data to train on, other than the official training dataset being competed on. In each case the open-source repository was first downloaded, and the original training protocol was run with recommended settings to determine a baseline. Once this baseline was calculated, Perforated Backpropagation was added into their pipeline and training was run again. Training proceeded in as many cycles as was required until early stopping when validation error stopped improving.

Our most impactful result in this study was on a network architecture called TrimNet~\cite{Li2021}. TrimNet is a graph neural network which utilizes a triplet message mechanism to learn molecular representation. We found TrimNet when it was the state-of-the-art system on the Tox 21 dataset from MoleculeNet~\cite{Wu2018}. Tox 21 is a toxicology dataset which is used to learn the toxicity of various chemical compounds. The dataset contains ``dense features'' that represent chemical descriptors and ``sparse features'' that represent chemical substructures. For each compound machine learning systems are expected to predict if the compound will be active or inactive in 12 different toxicological experiments.

\begin{figure*}[t]
\centering
\includegraphics[width=0.9\textwidth]{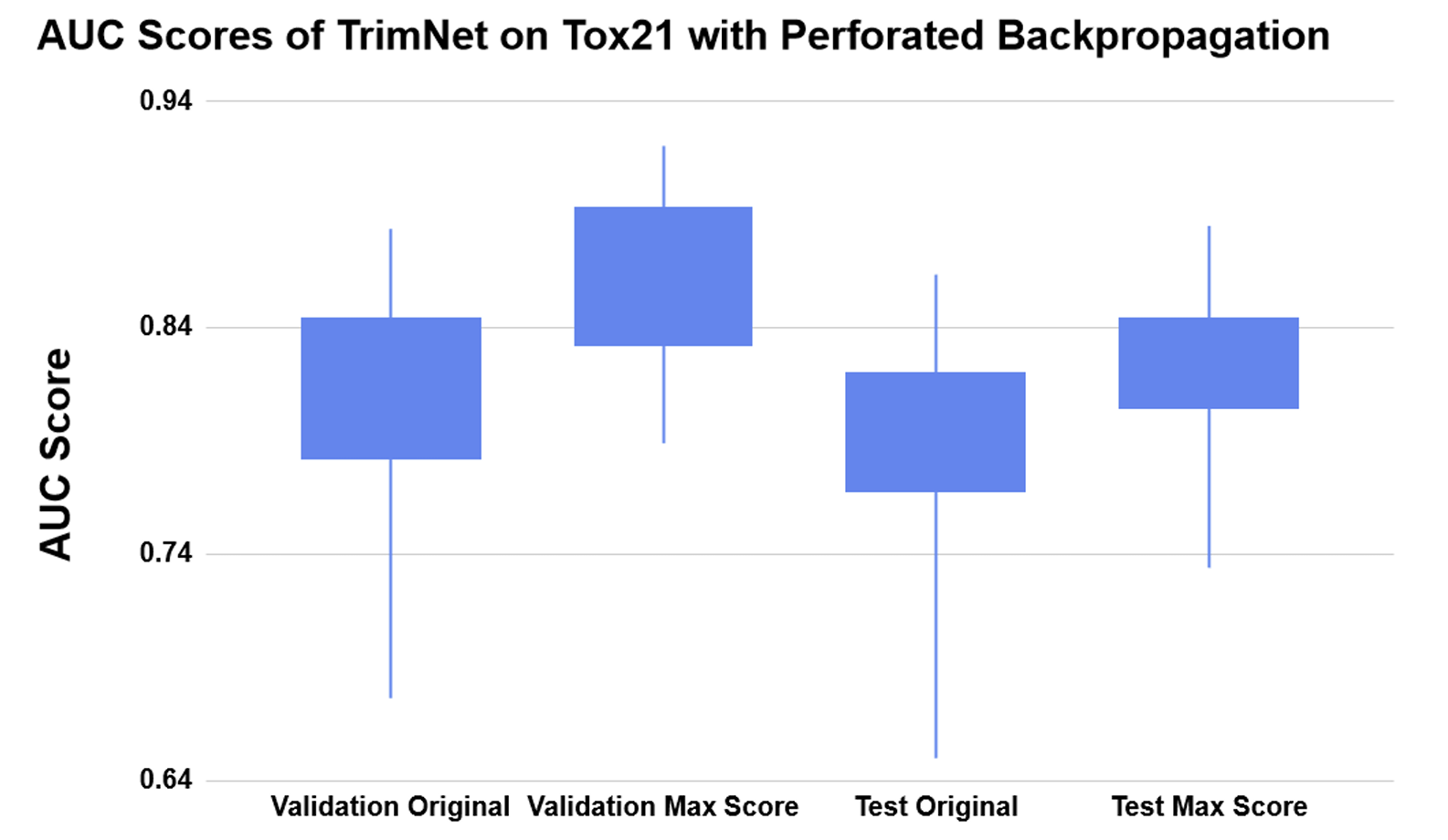}
\caption{Candlestick chart of AUC scores of our fifty runs of Perforated Backpropagation enhanced TrimNet on the Tox 21 dataset. Results show the max test score has higher values than the original test score.}
\label{fig:trimnet_auc}
\end{figure*}

The open-source code was downloaded and run with default settings to determine a baseline result. Perforated Backpropagation was then added to their system and the experiment was run fifty times with fifty different random initializations. This experiment was run over so many iterations because of the high fluctuation of scores depending on the random seed of the training. Over the experiments, on average, the test AUC score was able to be pushed 13.6\% closer to a perfect score of 1.0. The average AUC score when replicating the original experiment was 0.789. The average test score taken at the point of the highest validation score over a full experiment was 0.822. Training would be run until a cycle where additional dendrites started to decrease the validation score. A table showing how many dendrites were added to final models can be seen in Table~\ref{tab:dendrite_counts}. TrimNet's published result of 0.860 AUC is above our average, but below our highest test score over the experiments of 0.885. In that particular instance, our random initialization gave us a starting score of 0.807 which meant the error reduction of 40.4\% was much better than our average.

\begin{table*}[t]
\centering
\caption{Table showing how many dendrites were used for optimization in each of the fifty experiments. Two out of fifty experiments showed dendrites did not improve upon the original. The maximum added was a single experiment which added twelve.}
\label{tab:dendrite_counts}
\small
\setlength{\tabcolsep}{6pt}
\begin{tabular}{lccccccccccccc}
\toprule
Number of Dendrites & 0 & 1 & 2 & 3 & 4 & 5 & 6 & 7 & 8 & 9 & 10 & 11 & 12 \\
\midrule
Experiment Count & 2 & 2 & 8 & 11 & 5 & 2 & 7 & 7 & 2 & 2 & 1 & 0 & 1 \\
\bottomrule
\end{tabular}
\end{table*}

\begin{figure}[t]
\centering
\includegraphics[width=\columnwidth]{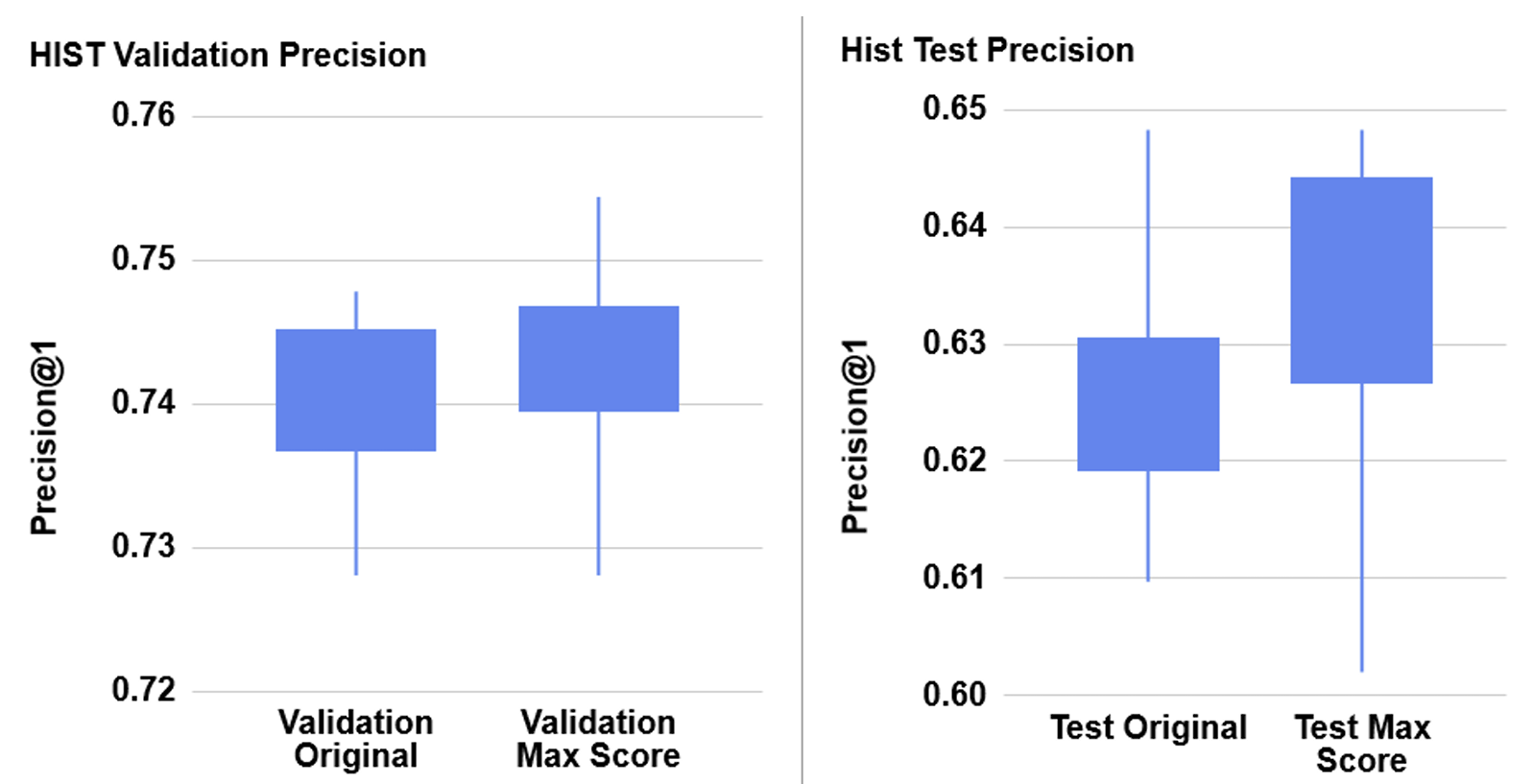}
\caption{Candlestick chart of validation and test scores of Perforated Backpropagation with HIST on the CSI300 dataset over ten experiments. Results show the max test score has higher values than the original test score.}
\label{fig:hist_scores}
\end{figure}

The second experiment was using Microsoft's qlib stock trend forecasting repository~\cite{qlib2023}. This repository contains a library of implemented models to train and test on the CSI300 dataset first introduced in~\cite{Han2023}. This dataset contains opening price, closing price, highest price, lowest price, volume weighted average price and volume looking back 60 days to construct a 360-dimensional historical stock data as input. Networks train on this data to make decisions about how each stock will perform in the future. The network chosen to work with was the HIST model~\cite{Xu2021HIST}. HIST is a graph-based framework designed to mine the concept-oriented shared information between stocks while making decisions about each stock. The Precision@1 metric is determined by the percentage of predictions where the top-rated stock has a correct label. Because random predictions would grant a score of 0.5, this was the score used as the baseline for determining our improvements above the original score. For example, our best result raised the original test score from 0.610 to 0.648, granting a 35\% increase above 0.5. On average, over 10 experiments with different random initializations an 8.6\% improvement was found.

\subsection{Test Addressing the Increase in Parameters}

Each time dendrites are added to these systems, when all modules are targeted, the total number of free parameters of the architecture increases by approximately the total number of parameters of the original model. This is because dendrites have weighted connections to each of the same input neurons as the neuron they are associated with. A question that this raises, is whether these networks are only improving because the total number of parameters is increasing. To answer that question, an experiment was run to determine the effect of parameter changes by adjusting the number of parameters with dendrites compared to adjusting the number of parameters via the width of the network. In a network with fully-connected layers, adjusting the number of nodes per layer causes a quadratic change in the number of parameters. Adding dendrites causes a linear change. This final experiment is on the PhysioNet Challenge from 2012~\cite{Silva2012}. The PhysioNet dataset is de-identified records of patients in Intensive Care Units. Each record consists of roughly 48 hours of irregularly sampled multivariate time series data used to predict in-hospital mortality of the patients. The system chosen to work with on this dataset is called the Multi Time Attention Network (mTAN)~\cite{Shukla2021}.

\begin{figure*}[t]
\centering
\includegraphics[width=0.9\textwidth]{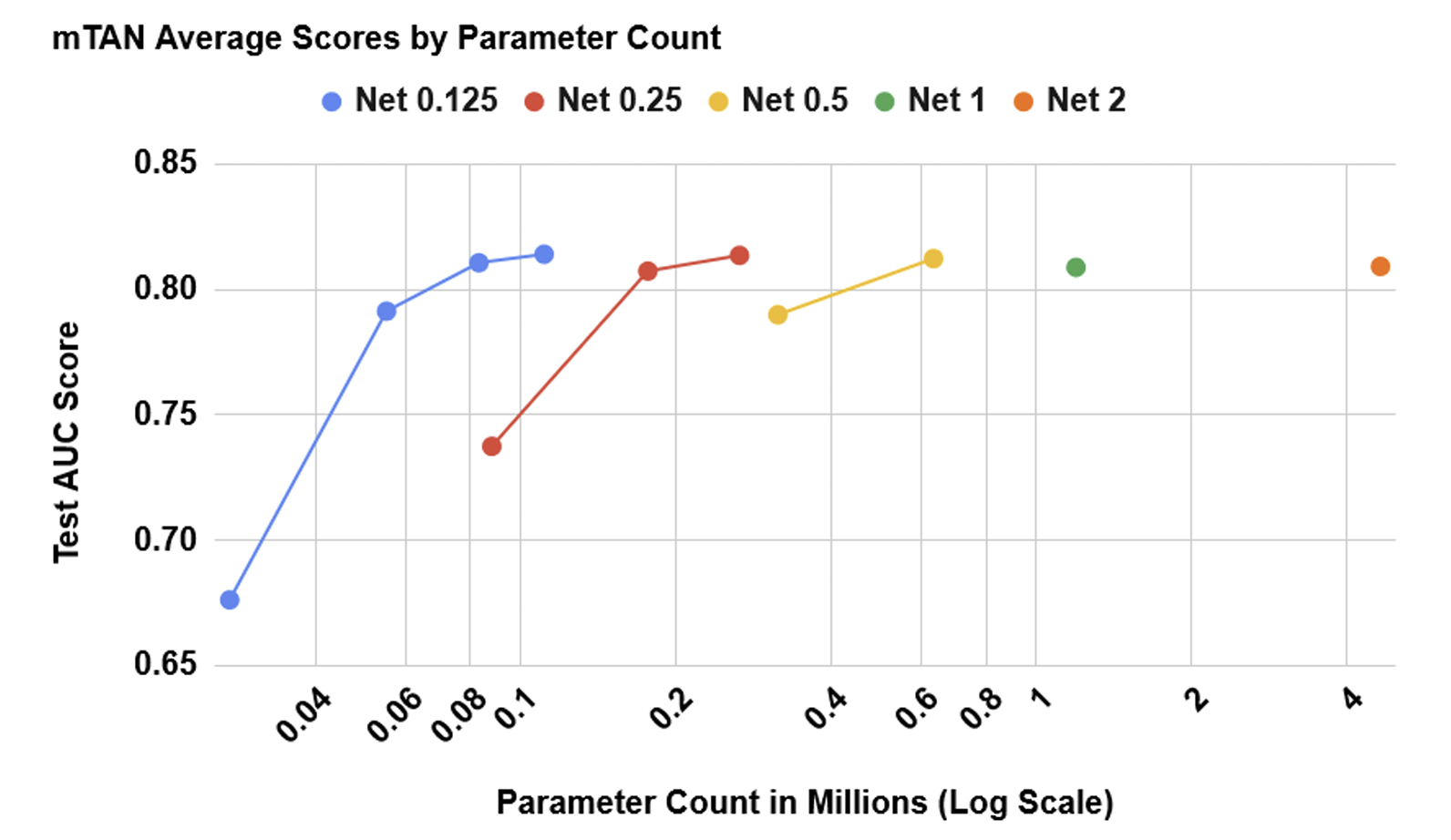}
\caption{Graph showing total parameter counts compared to average test AUC scores of each network tested over seven experiments. Networks are labeled by their width multiplier. Multiple points show network's initial values as well as subsequent values as dendrites are added. Results show that smaller initial networks with Perforated Backpropagation can create smaller final models without a loss in AUC.}
\label{fig:mtan_params}
\end{figure*}

A graph showing the results over various network widths can be seen in Figure~\ref{fig:mtan_params}. The leftmost point represents the original score of that architecture while each subsequent point shows the scores calculated with each additional dendrite added. Three dendrites were included for Net 0.125, two for Net 0.25 and one for Net 0.5 because there were rarely improvements with additional dendrites for each network after that number of cycles. Dendrites were not included for Net 1 and Net 2 because these networks regularly overfit with the addition of the first dendrite. In these experiments the final Net 0.125 architecture is not only less than a tenth the size of the original, but the average score of Net 0.125 with three dendrites is 2.7\% better than Net 1. Hence, our improved performance with dendrites cannot be explained by sheer parameter count.

In addition to dendrites impacting parameter counts, the architecture changes also cause increases to both training time and inference time. In this experiment both neuron training cycles and dendrite training cycles each consisted of 250 epochs. This means that while the original architecture only took 250 epochs in total, Net 0.125 trained for 1750 epochs. With Net 0.125's smaller size the original epochs were faster, but by the later epochs the additional dendrites caused each training epoch to be slower than the original. The average epoch time and total training times for each network can be seen in Table~\ref{tab:timing}. In addition, Table~\ref{tab:timing} contains timing information for running inference on each architecture. CPU inference was run a single batch at a time and GPU inference was processed in batches of 100. Inference times are not shown for dendrite cycles because inference is not run with networks which have dendrites that are not connected to neurons.

In this experiment an older implementation was used which caused dendrites to produce slowdowns at a higher rate such that a smaller perforated model ended up slower than a larger traditional model.  In the current library this is not the case, smaller perforated models are proportionally faster by parameter count than larger traditional models. Details of timing for the newer library will be included in ongoing work being prepared for publication.

\begin{table*}[t]
\centering
\caption{Timing information for training and inference of the mTAN network. Values show the addition of dendrites has a significant time cost during training and a modest time cost during inference.}
\label{tab:timing}
\small
\begin{tabular}{llrrrr}
\toprule
& & \begin{tabular}[c]{@{}r@{}}Seconds Per\\ Training Epoch\end{tabular}
  & \begin{tabular}[c]{@{}r@{}}Seconds Per\\ Training Cycle\end{tabular}
  & \begin{tabular}[c]{@{}r@{}}GPU Inference\\ Inputs/Second\end{tabular}
  & \begin{tabular}[c]{@{}r@{}}CPU Inference\\ Inputs/Second\end{tabular} \\
\midrule
\multirow{8}{*}{Net 0.125}
 & Neuron 1   & 5.15  & 1287.28 & 3026.21 & 52.12 \\
 & Neuron 2   & 7.59  & 1897.93 & 1528.8  & 30.24 \\
 & Neuron 3   & 10.03 & 2507.28 & 1019.72 & 21.90 \\
 & Neuron 4   & 12.61 & 3170.31 & 733.04  & 16.43 \\
 & Dendrite 1 & 9.43  & 2357.83 &         &       \\
 & Dendrite 2 & 11.89 & 2970.20 &         &       \\
 & Dendrite 3 & 14.68 & 3670.25 &         &       \\
 & Total Training Time & & 17861.08 & & \\
\midrule
\multirow{6}{*}{Net 0.25}
 & Neuron 1   & 5.20  & 1299.20 & 3059.39 & 41.31 \\
 & Neuron 2   & 7.59  & 1897.91 & 1551.21 & 25.60 \\
 & Neuron 3   & 9.98  & 2493.77 & 1030.59 & 17.52 \\
 & Dendrite 1 & 9.50  & 2375.40 &         &       \\
 & Dendrite 2 & 11.89 & 2972.27 &         &       \\
 & Total Training Time & & 11038.55 & & \\
\midrule
\multirow{4}{*}{Net 0.5}
 & Neuron 1   & 5.69  & 1423.27 & 2576.98 & 32.26 \\
 & Neuron 2   & 8.51  & 2126.69 & 1293.69 & 18.37 \\
 & Dendrite 1 & 10.70 & 2674.49 &         &       \\
 & Total Training Time & & 6224.45 & & \\
\midrule
Net 1 & Total Training Time & 7.13  & 1781.96 & 1783.37 & 26.46 \\
\midrule
Net 2 & Total Training Time & 10.41 & 2602.96 & 1177.45 & 15.90 \\
\bottomrule
\end{tabular}
\end{table*}

In neural network training there is a tradeoff between accuracy and model size. The first two experiments showed that dendrites can offer increased accuracy at the cost of increased parameters. This experiment shows that dendrites can also decrease parameters without a loss in accuracy. While inference only sees a modest increase in time to process each input, training time to add dendrites brings a significant increase in the total time to train a model because of how many additional epochs must be run.

\section{Possible Biological Implications}

Artificial neurons do not mirror the way biological neurons work. But when artificial neurons were first created, they were called neurons because they do share key features of biological neurons. Both combine input from presynaptic connections, the connections have the concept of a connection weight, and the input is passed through a nonlinearity after being combined. As with artificial neurons, the artificial dendrites of Perforated Backpropagation do not mirror the way biological dendrites work. However, it is equally appropriate to call these new nodes artificial dendrites as it is to call the original nodes artificial neurons because they share key features of biological dendrites which are not present in standard artificial neurons.

Two points regard the nodes themselves, while the rest are related to how they learn. First, the original artificial neurons simplified the complexity of a biological neuron's action potential with a simple thresholding function. This later evolved into sigmoid, TanH, ReLu and the other nonlinear functions more commonly found in modern networks. Similarly, these modern functions are used for dendrites rather than implementing the full complexity of a biological dendrite's method of nonlinear integration. Despite the simplifications of the true biological mechanics, we argue it is more biologically realistic to have presynaptic input for an artificial neuron go through additional nonlinear functions via dendrites rather than direct presynaptic-postsynaptic connections of traditional artificial neural networks. Second, dendrites are closer in architectural design to biological dendrites than they are to biological neurons, because they postsynaptically connect to only a single neuron and other dendrites connected to that neuron.

Biology also supports key features of our perforated learning paradigm. Hebbian learning~\cite{Hebb1949} is an accepted system in neuroscience, the idea that when neurons fire together temporally their connection is strengthened. Connection strengths are also modified with other mechanisms, many of which transpire on the postsynaptic side of the synapse. As some examples, postsynaptic cells exhibiting a rise in calcium can lead to increased neurotransmitter receptors at synapses~\cite{Gerrow2010}. To stabilize Hebbian learning average level of postsynaptic depolarization can decrease synapse strength~\cite{Leslie2001, Abbott2000}. Postsynaptic synapse size has been shown to be directly related to synapse strength~\cite{Meyer2014}. Protein kinase A, found more concentrated in dendrites than the rest of the neuron, regulates long term potentiation~\cite{Zhong2009}. This idea that so much of synapse strength is modified by the postsynaptic cell is evidence that in biology the post synaptic cell has great influence on the connection strengths it forms to presynaptic neurons. Perforated Backpropagation also has a learning rule where the error calculated at the postsynaptic artificial neuron is the value used to adjust the strengths of all connections into its dendrites.

We hope future experiments with this algorithm could also be shown to be more computationally biologically realistic than previous models have been. Much work has been done trying to prove that current instantiations of convolutional neural nets already provide a parallel to the biological brain~\cite{Parker2015, Kubilius2016, KhalighRazavi2014}. But there is significant evidence against CNNs learning neurologically relevant representations of the human visual system.

This evidence exists both on the level of individual neurons and networks as a whole. Research has shown that in biology individual neurons are sensitive to non-accidental property (NAP) changes in visual shape processing and therefore are coding for some shape aspect which is more invariant to metric changes than to NAP changes~\cite{Kayaert2003}. In artificial CNNs, neurons instead learn ``uninterpretable solutions'' and that the individual top layer units do not code for anything more specific than random linear combinations of top layer units~\cite{Goodfellow2009, Zeiler2014}. Other neural network papers often show various input images or pixel clusters that most strongly activate neurons to argue that each particular neuron is sensitive to a specific quality and claim it learned something important~\cite{Krizhevsky2012}. One group tested neurons at equal depths, from networks which had been trained and networks which instead had been given random weights. Two tests are performed, one on MNIST handwritten digits~\cite{LeCun1998} and a second on a deeper network against ImageNet images~\cite{Deng2009}. The results show that nodes with random weights show the same type of results that some researchers had claimed proves the neurons learned a useful semantic property contained in the training images. This experiment suggests that, when looking at the calculation of a single individual neuron in that network, a random set of weights could be interpreted to be equally useful as a trained node~\cite{Szegedy2013}.

More evidence that CNNs are not utilizing a neurologically relevant representation comes from viewing networks as a whole, testing what kind of input images can cause a fully trained deep CNN to make mistakes. Adversarial images can confuse artificial neural networks~\cite{Goodfellow2014}. If neurons coded for 3-dimensional structure through a hierarchy of NAPs to geons to objects there would be little difference in the top layer activations from these minor changes to the input. Results also showed across networks and training sets the same adversarial examples would cause similar mistakes. This suggests networks do learn a consistent representation of the input space~\cite{Szegedy2013}. Other research has shown CNNs even learn certain invariances to changes in input~\cite{Zeiler2014}. However, failing on these types of changes provides evidence that the representation and invariances learned are not the same as in the brain. For many of these examples a human viewer would find the modified images indistinguishable from their originals. One cannot argue that these deep networks are not doing impressive calculations, but this is strong evidence that they are not doing them in neurologically relevant ways. Performing similar experiments with our system is planned for future work.

\section{Conclusion}

This paper has shown a novel extension to artificial neural network learning. Based on biological dendrites' non-linear spiking, new nodes can be added to artificial networks. These nodes are used to modify the way presynaptic input is processed to empower single neurons to have improved performance coding for the same features they were originally. Learning with Perforated Backpropagation, where the error gradient does not pass through every node in the network, these new nodes learn input weights to only help reduce error for a single node in the network. Using this framework our results presented here show that whether you are starting with a multi-layer perceptron, a deep convolutional neural network, or a graph network once error has reached a threshold it can be reduced further by adding dendrites. Future works being prepared for publication show that this expands to transformer architectures, tabular models, and many more architecture families.  Our framework also allows for significant model compression by starting with a reduced framework and then adding dendrites to reach the original accuracy while maintaining a reduced size.

These results are highly encouraging that artificial dendrites performing additional non-linear processing for individual neurons in between hidden layers of deep neural networks could be a promising direction to create better machine learning systems. We publish this research with the hope that there will be interest in working with our approach in the deep learning community.

\subsection{Reproducing Results and Perforating Your Own Models}

Instructions and best practices to add Perforated Backpropagation to any project built with PyTorch, including agent skills to perforate a model with a single prompt, can be found in our main GitHub repository at \url{https://github.com/PerforatedAI/PerforatedAI}. Instructions to reproduce the results of this paper can be found in our secondary examples GitHub repository at \url{https://github.com/PerforatedAI/PerforatedAI-Examples}.

\bibliographystyle{ieeetr}
\bibliography{references}

\end{document}